# Real Time Fabric Defect Detection System on an Embedded DSP Platform

Jagdish Lal Raheja[1], Bandla Ajay[2], Ankit Chaudhary[3]

[1] Machine Vision Lab
CEERI/CSIR
Pilani, RJ India
jagdish@ceeri.ernet.in

[2]Dept. of Electrical & Electronics Engg.
BITS, Pilani, RJ India
ajayb5632@gmail.com

[3]Dept. of Electrical & Computer Engg.
The University of Iowa, USA
ankit-chaudhary@uiowa.edu

**Abstract-** In industrial fabric productions, automated real time systems are needed to find out the minor defects. It will save the cost by not transporting defected products and also would help in making compmay image of quality fabrics by sending out only undefected products. A real time fabric defect detection system (FDDS), implementd on an embedded DSP platform is presented here. Textural features of fabric image are extracted based on gray level co-occurrence matrix (GLCM). A sliding window technique is used for defect detection where window moves over the whole image computing a textural energy from the GLCM of the fabric image. The energy values are compared to a reference and the deviations beyond a threshold are reported as defects and also visually represented by a window. The implementation is carried out on a TI TMS320DM642 platform and programmed using code composer studio software. The real time output of this implementation was shown on a monitor.

**Keywords**- Fabric Defects, Texture, Grey Level Co-occurrence Matrix, DSP Kit, Energy Computation, Sliding Window, FDDS

## 1. Introduction

Identifying defects in fabric is a major concern for fabric industries as it is important for their brand images for quality products. Automated visual inspection system to detect possible defects in fabric provides a more reliable and consistent quality control process than human eye view. Many research work has been demonstrated in this area but a robust real time system for color fabrics is still needed. Efficient algorithms are also needed to reduce the response time in real time. Karayiannis [1] wrote for the need of a real time system about fifteen years back, "In the best case, a human can detect not more that the 60% of the present defects and he cannot deal with fabric wider than 2 meters and moving faster than 30 meters/min". However, implementing automated visual inspection has to meet with some challenges. A number of techniques for describing image texture have been proposed in the research literature. A good review on these techniques is described in [2].

There are a variety of defects occurring in fabrics, in terms of texture pattern violation, weaving defects, yarn misplacements etc. Tuceryan [3] defined five major categories of features for texture analysis: statistical, geometrical, structural, model based and spectral features. Spectral features are based on the periodicity of the texture feature. The high degree of periodicity of basic texture primitives, such as yarns in the case of textile fabric, permits the usage of spectral features for the detection of defects. Also, Chan [4] has been able to identify defects on simulated, real fabrics and then could classify these defects based on some extracted parameters. To perform real time critical operations for an industrial application, it require specialized hardware in operations, to ensure faster decision making time.

The wide variety of defects and textures encountered in fabric industries are difficult to identify by a fully automated visual inspection system. This paper limits the variety of fabric defects to some simple ones as those described by [5-6]. This paper is organized as follows. In the next section the background of the application and methods are discussed. A brief introduction of GLCM is shown in section 3 and mathematical model for fabric detection is described in section 4. Hardware and software components used for our systems are detailed in section 5 and 6 respectively. Later section 7 discuss obtained results followed by conclusions.

## 2. Background

Application of co-occurrence texture features for interpreting fabric defects are discussed in this section. Many research work could be found on real time fabric defect implementation using different methods for real time applications. Guang [7] presents a defect detection algorithm based on distance calculation among images where he claims accracy of 92%. Han [8] use gabor filter with GA to detect fabric defects. Shu [9] detailed a method of detecting the fabric defects automatically based on multi-channel and multi-scale Gabor filtering. It is based on energy response from the convolution of gabor filter banks in different frequency and orientation domains. These sets of banks are to be convolved with the image to generate output as a set of images. Further, these images need to undergo smart segmentation processing to identify defective regions. After this, these images should be under a decision making system which decides for defects in the image. The implementation using gabor filter banks is computationally very expensive. These filters are 16x16 arrays, which need to be convolved in two dimensional spaces with the image.

Karayiannis [1] presented multiresolution decomposition based real time FDDS. He used AT&T 32C DSP board for implementation while a pentium PC was used with PC bus to show output. Jia [10] presented a edge based seed filling algorithm for fabric detection. Different edge detections methods with connecting broken edges have been discussed in [11]. Shi [12] used a segmentation method based on local contrast deviation for fabric defect detection. Real time region of interst croping for faster processing is discussed in [13]. Zhang [14-15] presented a multiple window gray ratio based algorihtm for cord fabric detection. Although it work in real time but cord is a different kind of fabric and he used gray images to find the defect.

Siew [16] has shown assessment of carpet wear using spatial gray level dependence matrix (SGLDM). Also methods similar to GLCM have been applied to wood inspection [17], surface defect detection [18], and fabric defect detection [19]. The original investigation into SGLDM features was pioneered by Harlick [20]. Texture features such as energy, entropy, contrast, homogeneity, and correlation are then derived from the co-occurrence matrix. However only six of such features have been used for the defect detection on wood and fabric defect detection has been shown with only two of these six features. Interestingly Conners [17] has used six features of co-occurrence matrix, to identify nine different kinds of surface defect in wood. Also, Tsai [19] has detailed fabric defect detection while using only two features and achieved a classification rate as high as 96%. Zhang [21] came with synthetic aperture radar to overcome disadvantages of brightness effect of light at the time of defect detection but results were not so clear.

As far as real time implementation is concerned, Bariamis [22] has proposed an FPGA based architecture for real time image feature extraction, which describes a FPGA based system to compute GLCM. Mak [23] used iterative tensor tracking using spatial histogram [24] of gradient orientations on a GPU NVIDIA GTX 260 O.C.; He showed good compuatation results as the image resolution increases.

## 3. Gray Level Co-occurrence Matrix

The GLCM method, also known as the spatial gray-level dependence method, has been widely used in texture analysis, as discussed in previous section. It is based on repeated occurrences of different grey level configurations in a texture. Automatic visual inspection techniques for textured images generally compute a set of textural features in the spatial domain or in the spectral domain. The co-occurrence probabilities provide a second order method for generating texture features. A brief presentation of the GLCM method follows but a more complete explanation is provided by Haralick [20][25].

This GLCM matrix contains the conditional joint probabilities of all pair wise combinations of grey levels given two parameter: inter-pixel distance (say d) and inter-pixel orientation (say θ). They were measured counter clockwise from the horizontal axis. The probability measure is defined by Barber [26] in terms of number of occurrences of gray levels within a region for a particular value of displacement and orientation. Different statistical information can be determined from each GLCM. However, statistics that are grey level shift invariant are important so that the classification is not a function of tone. Eight such shift invariant statistics are presented in Haralick's work [25]. Energy (ENG), Entropy (ENT) and maximum probability (MAX) are linear (scale and shift) invariant statistics.

## 4. Defect detection model using GLCM

The system architecture was designed on the principle of components [27]. There are a number of different parameters that must be indicated in order to generate co-occurrence data: window size (say n), orientation (θ), pixel separation distance (d) and the number of quantized grey levels (G). This paper proposes a defect detection method where a moving window, slides over the whole image in steps computing GLCM at each position. Then energy feature is extracted from the GLCM. These energy values are compared to that of a reference and then any deviation beyond a threshold is identified as defect.

For classification studies, the window size is often fixed and assumes that sufficient area of the texture is represented to capture an appropriate measurement. Typically θ is set to a value among 0, 45, 90 and 135 degrees, since this is easiest to implement. To program the generation of co-occurrence data for arbitrary orientations is not reasonable. Short pixel separation distances have typically achieved the best success [21]. Various values of d can yield different features. However, it is possible if there is any priori method for selecting its value based on the given image characteristics.

Here for defect detection, parameters were choosed as d=1, theta=0, G=256 (for 8 bit image) and window size to 50x50. Haralick [25] introduced several texture features on GLCM where we use energy feature to identify defects. The minimum window size should be such that it should include the at least one periodic pattern of the texture. The fabric should have a repeating pattern of the image in the window as. A larger window which includes three or more periodic patterns will identify the defects but it will fail to properly localize the defective region in the fabric. That is, the whole window region will be shown as defective in the event of an observable difference in energy. Although a larger window scans the image in less number of steps, it takes more time to perform extraction operation of the region of interest. The number of computations for a single image is as follows

1) Number of steps to scan the whole image = [R/w] * [C/w]     ---- (1)

Where R and C are the number of rows and columns in image and w is the window size. '[ ]' denotes the floor value.

2) Number of computations at each position of the window:

  2.1) to extract the window region from the input image    = w * w    ---- (2)
  2.2) to compute GLCM    = G * G    ---- (3)

So the total number of steps required to identify defects in an image of size RxC is

$$[R/w]*[C/w]*w*w*G*G = R*C*G*G \quad ---- (4)$$

Table 1 shows the number of computations for an image of dimensions R= 461, C= 512 and G= 256 (8 bit gray scale).

Table 1: Total number of Computations

| Window size | No of steps required to slide through the whole image[+] | Total no of computations |
|---|---|---|
| 30 | 255 | 15e9 |
| 50 | 90 | 14e9 |
| 100 | 20 | 13e9 |

[+]A count variable keeps track of this number in the program.

Hence, the larger windows slightly reduce the numbers of computations. The window size should be chosen to balance the two factors i.e. localizing the defect and reducing computations. Initially to identify important textural features, the deviations of energy along with other features were noted as a window moved over the image.

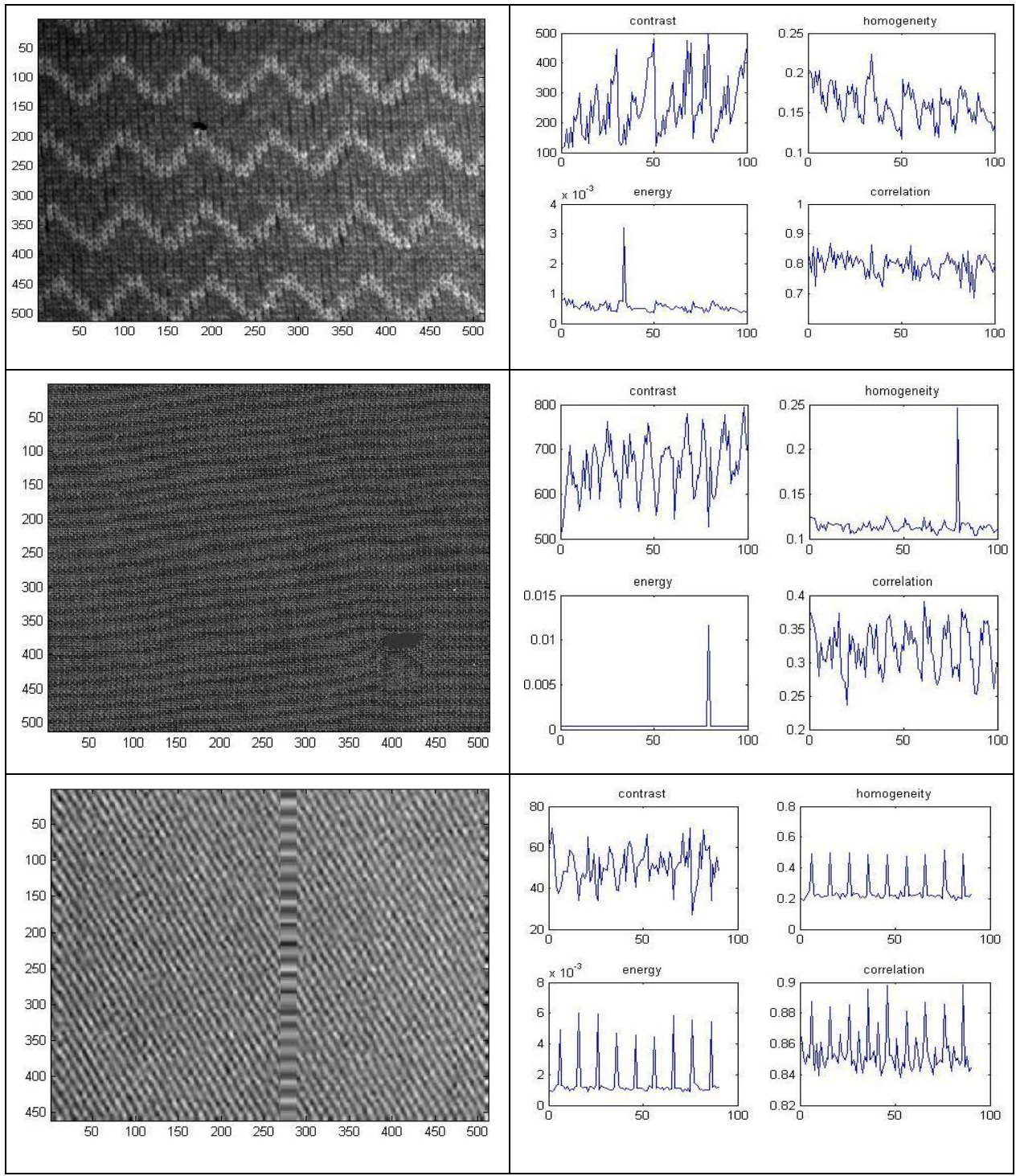

Figure 1: Images on the left are defective image samples, images on the right are graphs of the features (1.Contrast, 2. Homogeneity, 3. Energy, and 4.Correlation) extracted from the sample image as a 50x50 window slides over the whole image. The x axis denotes the count of movement of the window as it slides through the whole image. The window moves in steps of 50 pixels.

The moving window type defect detection logic using energy, correlation homogeneity and contrast was also tested. Out of these features, energy feature always gave prominent peaks in the defective regions compared to the remaining textural features. This was tested for several defective images, which also showed observable peaks for energy feature. Hence this energy feature was used to identify texture defects in the image. Figure 1 shows samples and their graph of their features. It concludes that energy feature contains information to identify the defects in the image. The defect detection results of this logic, using energy feature are implemented in MATLAB. Figure 2 shows few results of defect detection.

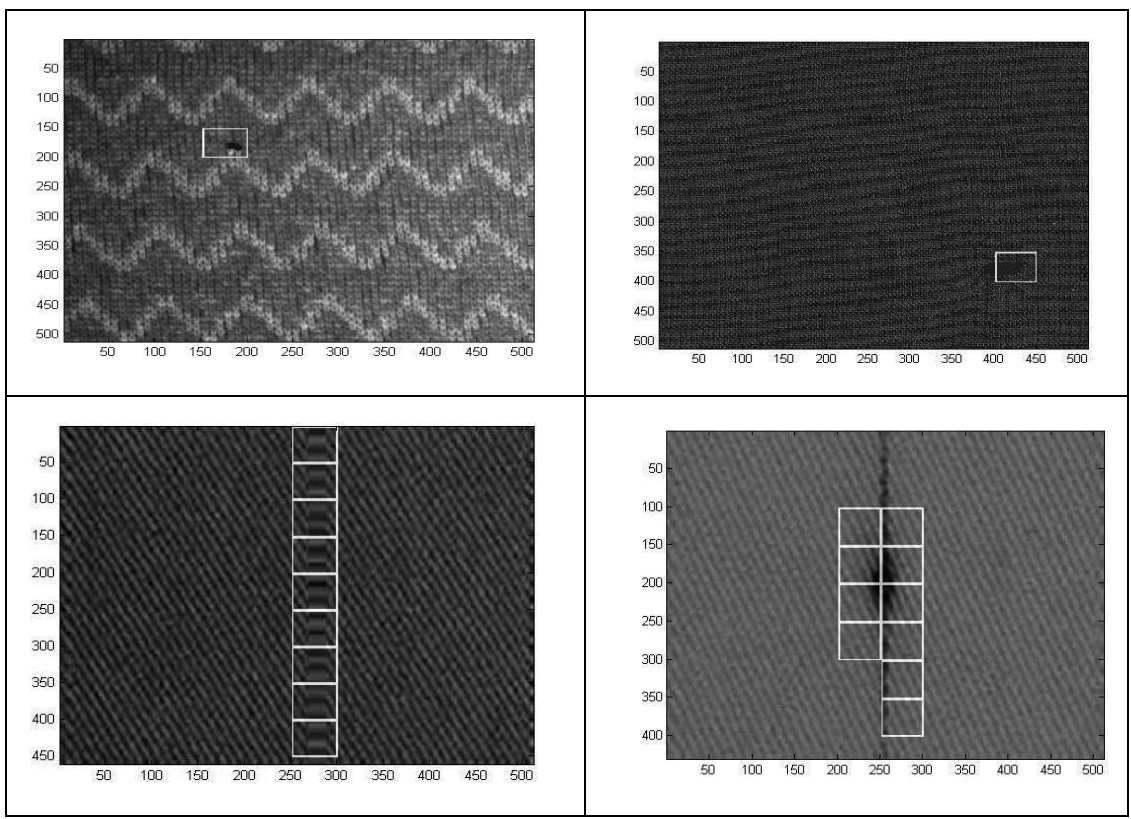

Figure 2: Defect detection results of moving window implementation by using energy feature for texture defect detection.

## 5. Hardware

The FDDS was implemented on a Texas Instruments' TMS320DM642 Evaluation Module. It has three on board video ports with 2 decoders and 1 encoder. The Input video was taken from a digital color camera in NTSC format. The gray level information was extracted from one of the channels. The direct memory access (DMA) feature of the DSP kit facilitates memory to memory transfer operations without disturbing the processor operation cycle. The image was loaded to the memory from SDRAM through this DMA feature. Image processing functions were performed and the final output with markings of defects if any, is transferred to the output buffer. This code was tested for real time operations by taking the input from a real fabric with some defects as described in [5-6]. Figure 3 shows the algorithmic flow of the proposed approach.

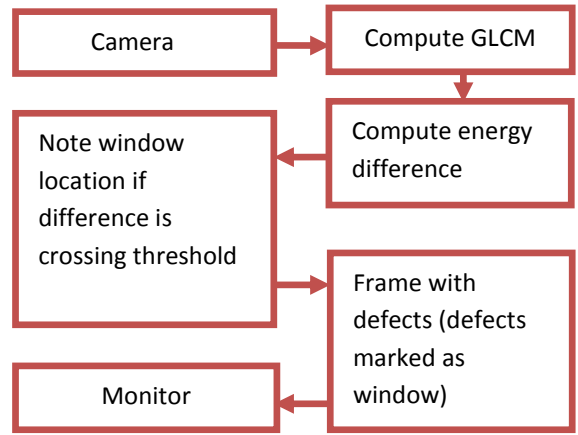

Figure 3: Flow of control in the program.

## 6. Software

This paper uses code composer studio IDE with additional libraries which come along with the Texas Instruments DSP kit to implement FDDS. This Texas Instruments Chip support Libraries (CSL) APIs [29] were used for configuring and controlling the DSP on-chip peripherals. The proposed logic was implemented after testing it in MATLAB[®] [28] environment running on Microsoft Windows[®]. The implementation program in code composer studio was written in C language. The libraries in code composer which have implemented certain register level operations which the user can not do explicitly. It also has some library functions performing major chip level operations which are named under CSL. It also has predefined video/image processing libraries which reduce the program development time for the user. The software also provide supports to perform functions relating to video port handles, exchange etc.

After the code has been developed, the program was loaded on to the DSP kit. The program includes some necessary information about location of memory segment labels, location of main, heap location etc. Once the program is loaded on to the chip it runs on the DSP kit performing the implemented program. More information about programming using Code Composer Studio can be found in the manuals [29-30] which come with the DSP kit.

## 7. Results

FDDS implemented on a Texas Instruments TMS320DM642 Evaluation Module, process one frame at a time. Moving window analysis is performed on each frame and the defects are marked with white squares on the input fabric image. This image with defective regions marked as white squares is shown on a monitor connected to the DSP kit. The camera was positioned to capture 4 cm$^2$ area of the fabric in each frame. The response time of results on the monitior for the frames processed is 0.5 seconds. In laboratory setuo, the fabric was manually moved to get the next focus area. The technique is still primitive in terms of setting up motors to keep the fabric rolling, moving the camera etc. which are required for a full scale industrial use. The results of FDDS on the monitor are shown in Figure 4.

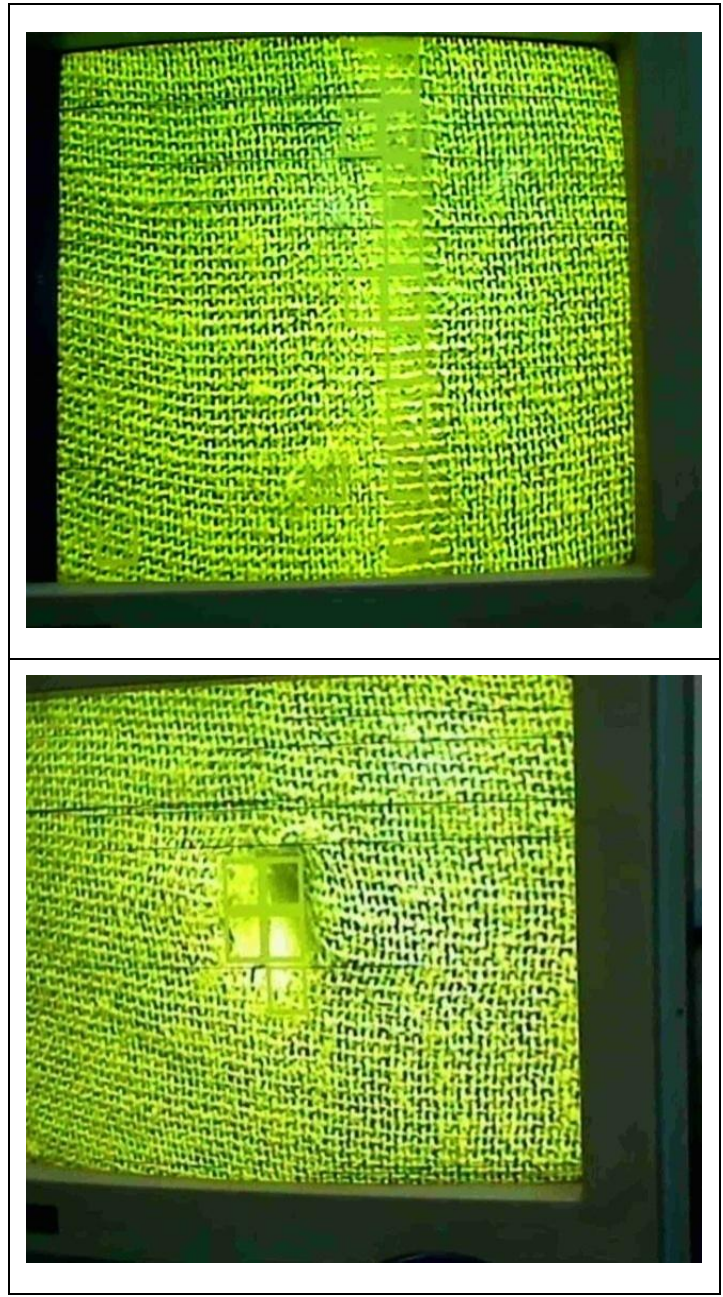

Figure 4: FDDS results on the Monitor.

## 8. Conclusions

The implemention of an automated visual inspection system for fabric defect detection is a challenging task. Most of the defect detection algorithms implemented previously in this area require some modifications in the program to properly identify the defects in case a new texture is needed to process. This paper implements a real time FDDS on a Texas Instruments DSP kit with input taken from a piece of fabric. The properties to identify the defects were tested using MATLAB environment. The program is fully automated and don't need any manual interaction once started. The results show that energy feature would be best to use to get information to distinctly identify defects in fabric textures. The results shown in this paper clearly identify the parameter to detect defects in textures. Also, the algorithm is simple in its decision making process and can be modified for extention. The response time is quite good and easily implimented in industrial applications.

## Acknowledgment

Authors would like to thank the Director of Central Electronic Engineering Research Institute/Council of Scientific and Industrial Research (CEERI/CSIR), Pilani, for providing research facilities, and for his active encouragement and support.